\documentclass[10pt,twocolumn,letterpaper]{article}

\usepackage{iccv}
\usepackage{times}
\usepackage{epsfig}
\usepackage{graphicx}
\usepackage{amsmath}
\usepackage{amssymb}
\usepackage{bbm}

\usepackage{rotating}

\DeclareMathOperator*{\argmin}{arg\,min}
\usepackage{booktabs}
\usepackage{multirow}
\usepackage{makecell}

\graphicspath{ {images/} }

\usepackage{amssymb}
\usepackage{pifont}
\newcommand{\cmark}{\ding{51}}
\newcommand{\xmark}{\ding{55}}

\usepackage[dvipsnames]{xcolor}

\newcommand{\Model}{DenoiseLoc}
\newcommand{\model}{DenoiseLoc}

\usepackage[pagebackref=true,breaklinks=true,letterpaper=true,colorlinks,bookmarks=false]{hyperref}

\iccvfinalcopy 

\def\iccvPaperID{****}

\ificcvfinal\fi

\begin{document}

\title{Boundary-Denoising for Video Activity Localization}

{\author{
Mengmeng Xu$^{1}$
\quad\quad\quad Mattia Soldan$^{1}$
\quad\quad\quad Jialin Gao$^{2}$
\quad\quad\quad Shuming Liu$^{1}$ \\
\quad\quad\quad Juan-Manuel P\'erez-R\'ua$^{3}$
\quad\quad\quad Bernard Ghanem$^{1}$
\and
{\small$^{1}$KAUST, KSA \quad\quad\quad $^{2}$NUS, Singapore \quad\quad\quad $^{3}$Meta AI}\\
}}
\maketitle

\maketitle

\ificcvfinal\thispagestyle{empty}\fi

\begin{abstract}


Video activity localization aims at understanding the semantic content in long untrimmed videos and retrieving actions of interest. The retrieved action with its start and end locations can be used for highlight generation, temporal action detection, etc. Unfortunately, learning the exact boundary location of activities is highly challenging because temporal activities are continuous in time, and there are often no clear-cut transitions between actions. Moreover, the definition of the start and end of events is subjective, which may confuse the model. To alleviate the boundary ambiguity, we propose to study the video activity localization problem from a denoising perspective. Specifically, we propose an encoder-decoder model named \model. During training, a set of action spans is randomly generated from the ground truth with a controlled noise scale. Then we attempt to reverse this process by boundary denoising, allowing the localizer to predict activities with precise boundaries and resulting in faster convergence speed. Experiments show that \model~ advances 
several video activity understanding tasks. For example, we observe a gain of +12.36\% average mAP on QV-Highlights dataset and +1.64\% mAP@0.5 on THUMOS'14 dataset over the baseline. Moreover, \model~achieves state-of-the-art performance on TACoS and MAD datasets, but with much fewer predictions compared to other current methods. 
\end{abstract}
\section{Introduction}
\label{sec:introduction}
\begin{figure}[ht]
\centering
\includegraphics[trim={0cm 0cm 0cm 0cm},width=.9\linewidth,clip,]{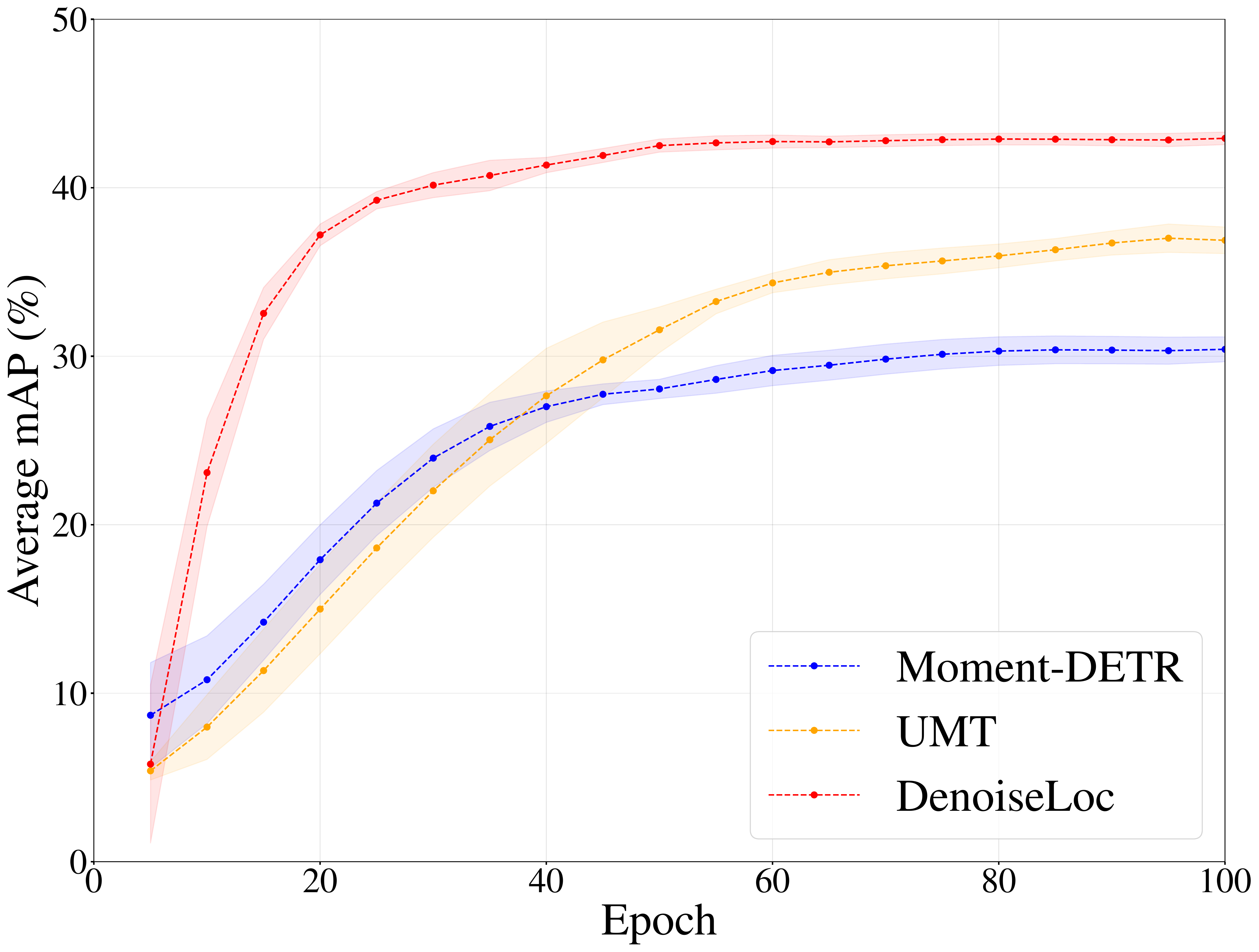}
\caption{DenoiseLoc is an encoder-decoder model that incorporates a temporal span denoising task during training, resulting in faster convergence and better performance compared to other state-of-the-art methods on the QV-Highlights dataset.}
\label{fig:pool}
\vspace{-.3cm}
\end{figure}

The summation of human experience is being expanded at a prodigious rate, making it imperative to devise efficient and effective information retrieval systems to aid the need for knowledge abstraction and dissemination. 
Recently, video data has emerged as the largest unstructured knowledge repository \cite{kay2017kinetics, ran2018deepdecision}, making it essential to develop algorithms that can understand and identify semantically relevant information within videos \cite{wu2019long, wu2017deep}. 
Our research focuses on the video activity localization domain \cite{gao2017tall,zhao2017temporal}, which enables users to identify, classify, and retrieve interesting video moments. Video activity localization tasks are defined to predict a set of temporal spans relevant to either a fixed class taxonomy \cite{caba2015activitynet} or free-form natural language queries \cite{he2019read}. These algorithms have numerous applications, including highlight generation \cite{lei2021detecting}, product placement \cite{jiao2021new}, and video editing \cite{neimark2021video}.

Technical solutions for activity localization often draw inspiration from innovations in object detection, which is a well-studied problem in the image domain. An analogy can, in fact, be drawn between spatially localizing objects and temporally localizing moments, with the key difference being that temporal boundaries are subjective and open to interpretation. 
A prevailing approach for model design in object detection is to adopt an encoder-decoder design paired with a suitable training protocol~\cite{carion2020end, moment-detr}.   
For video activity localization, the encoder processes the raw video frames and, optionally, a language query, with the goal of generating rich semantic representations that capture the interaction within the video and/or between video and language. These representations are referred to as ``memory''. The decoder leverages the encoder's memory to produce a list of temporal locations with corresponding confidence scores. The decoder achieves this by inputting candidate spans, which can be either predefined based on ground truth activity locations statistics or learned during the training process.

The primary challenge in video localization tasks stems from the \textbf{\textit{boundary ambiguity}} mentioned earlier. Unlike object boundaries, activities are continuous in time, and the saliency of a temporal event changes smoothly due to its non-zero momentum. 
Thus, transitions between activities are not always clear or intuitive.
Moreover, human perception of action boundaries is instinctive and subjective. This phenomenon is reflected in the existing video datasets, where we can identify multiple clues indicating that the variance of localization information is higher than for object locations. 
To support this thesis, DETAD~\cite{alwassel2018diagnosing} conducted a campaign to re-annotate the ActivityNet \cite{caba2015activitynet} dataset to estimate the annotator's agreement and quantify the severity of boundary uncertainty in humans. 
Additionally, a recent study~\cite{lei2021detecting} found that over 10\% of the queries in QV-Highlights showed disagreement on the boundary location with an Intersection over Union (IoU) of $0.9$. Thus, addressing the uncertainty of action boundaries is crucial for developing reliable and accurate video localization pipelines.

In this study, we aim to address the challenge of uncertain action boundaries in video activity localization. 
To this end, we propose \model, an encoder-decoder model which introduces a novel \textbf{boundary-denoising training} paradigm. 
In \model, the transformer-based encoder captures the relations within and across modalities. 
The decoder is provided with learnable proposals and noisy ground truth spans and progressively refines them across multiple decoder layers. In detail, we iteratively extract location-sensitive features and use them to update the proposal embeddings and spans in each decoder layer. Our boundary-denoising training jitters action proposals and serves as an augmentation to guide the model on predicting meaningful boundaries under the uncertainty of initial noisy spans.
Our denoising training method is easily generalizable to several video domains (i.e., YouTube videos, movies, egocentric videos) as it does not require hand-crafted proposal designs. Surprisingly, we find that with very few proposals (i.e., 30 per video), our model retrieves sufficient high-quality actions and performs on par with or better than computationally expensive methods evaluating thousands of proposals. Furthermore, we demonstrate that our denoising training yields a fast converging model, as shown in Fig.~\ref{fig:pool}.

To demonstrate the effectiveness and generalization ability of our proposed model, we conducted experiments on multiple localization tasks and datasets. Our results show that \model~achieved state-of-the-art performance on the MAD~\cite{Soldan_2022_CVPR} and QV-Highlights~\cite{lei2021detecting} datasets with significant improvements of 2.69\% on Recall@1 and 6.84\% on average mAP respectively. Moreover, \model~obtained modest performances on the TACoS~\cite{TACoS_ACL_2013} and THUMOS~\cite{jiang2014thumos} datasets. 

Our contributions are:
\textbf{(1)} A novel boundary-denoising training approach for video activity localization tasks. The noise injected at training time leads to faster convergence with a much smaller number of queries. 
\textbf{(2)} \model, an encoder-decoder style multi-modality localization model, specifically designed to exploit the boundary-denoising training strategy.
\textbf{(3)} Extensive experiments to demonstrate the effectiveness of \model, achieving state-of-the-art performance on several datasets, including the large-scale MAD dataset. Additionally, we provide thorough ablation and analysis studies to investigate the design choices and modeling properties of our approach.


\section{Related work}
\label{sec:related}
\begin{figure*}[ht]
\vspace{-0.5cm}
\centering
\includegraphics[trim={0cm 0cm 0cm 0cm},width=.9\textwidth,clip,]{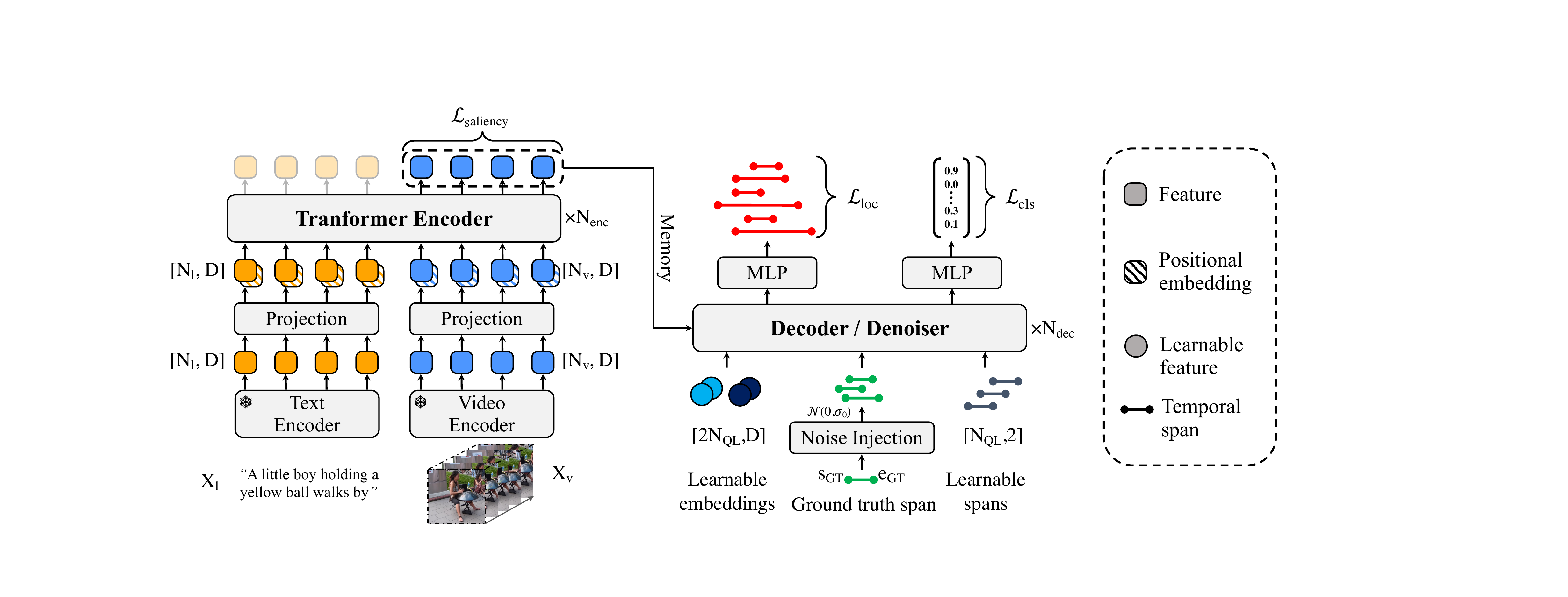}

\caption{\textbf{Architecture.} Our proposed model consists of three main modules: an encoder, a decoder, and a denoiser. 
First, the video features are fed into the encoder, and when available, the language query embeddings are concatenated with the video feature after a projection layer. Then, the output of the encoder is forwarded to a decoder and a denoiser. 
The decoder applies instance feature extraction and instance-instance interaction, while the denoiser module shares weights with the decoder but learns to refine an additional set of proposals, which are derived from the ground truth spans with a controlled injection of noise.}

\label{fig:arch}
\end{figure*}

\noindent\textbf{Natural language video grounding.}
Natural language video grounding is the task of predicting the temporal span in a video corresponding to a human-generated natural language description. Many solutions have been proposed for this task; however, methods can be clustered into two groups: \textbf{(i)} proposal-based methods~\cite{Gao_2017_ICCV,Hendricks_2017_ICCV,soldan2021vlg, 2DTAN_2020_AAAI, escorcia2019temporal}, which produce confidence or alignment scores for a predefined set of $M$ temporal moments and \textbf{(ii)} proposal-free~\cite{moment-detr, Zeng_2020_CVPR, Mun_2020_CVPR, chenhierarchical, Rodriguez_2020_WACV, Li_Guo_Wang_2021}, which directly regress the temporal interval boundaries for a given video-query pair. 
Proposal-based methods tend to outperform regression-based methods, but the latter have much faster inference times since they don't require exhaustive matching to a set of proposals. 
Our proposed \model~follows the proposal-free pipeline, and it is evaluated our model on both short-form (QV-Highlights~\cite{lei2021detecting} and TACoS~\cite{TACoS_ACL_2013}) and long-form (MAD~\cite{Soldan_2022_CVPR}) grounding datasets.

\noindent\textbf{Temporal activity localization.}  
Temporal activity localization (TAL) involves identifying actions and their categories within a video \cite{shou2017cdc, lin2018bsn, lin2019bmn, xu2020g}. Therefore, a language video grounding method can also be applied to TAL task if we omit the text prompt and classify the proposal category. In this work, we apply our localization model to TAL and demonstrate that our proposed denoising training technique leads to more accurate action predictions.


\noindent\textbf{Denoising and Diffusion.}
\label{sec:related-denoising-diffusion}
The issue of boundary noise has been extensively studied in the field of weakly supervised temporal action localization, where temporal annotation information is unavailable. To mitigate noise in generated pseudo labels, Yang, \etal~\cite{yang2021uncertainty,yang2022uncertainty}  designed an uncertainty-aware learning module UGCT, while Li et al.\cite{li2022exploring} proposed a novel Denoised Cross-video Contrastive (DCC) algorithm to reduce the negative impacts of noisy contrastive features. In fully supervised localization, Huang \etal~\cite{huang2022video} introduces Elastic Moment Bounding (EMB) to accommodate flexible and adaptive activity temporal boundaries with tolerance to underlying temporal uncertainties. However, these studies mainly focus on noise modeling rather than denoising.

More recently, object detection studies have explored learning with noise. DN-DETR~\cite{li2022dn} discovers that including noisy object proposals can accelerate the training of transformer-based detectors such as DETR~\cite{carion2020end}. 
DINO~\cite{zhang2022dino} proposes to add noise to ground truth at different levels to generate positive and negative samples for training. Meanwhile, DiffusionDet~\cite{chen2022diffusiondet} proposes a scheduled iterative denoising method to train a model to denoise the input proposals with different noise levels.  
Although our work is inspired by these methods, denoising on the temporal dimension is more challenging due to the fewer distinguishable features for neighboring snippets and a smaller number of instances as positives during training.

\section{Method}
\label{sec:method}

\subsection{Problem formulation}
\label{sec:method-problem}
Our task is to localize temporal instances that are relevant to either a pre-defined activity list or a natural language query.
The input video is modeled as a sequence of $n_v$ snippets, $V{=}\{v_i\}_1^{n_v}$, each comprising $\epsilon$ consecutive frames.
If available, the language query is tokenized in $n_l$ elements $L{=}\{l_i\}_1^{n_l}$.
Both inputs are mapped to multidimensional feature vectors using pre-trained models and are identified as $X_v \in \mathcal{R}^{c_v\times n_v}$,$X_l \in \mathcal{R}^{c_l\times n_l}$, respectively, where $c_v$ and $c_l$ are the feature dimension of snippets and tokens. 

To evaluate our model, we compare its predictions $\hat{\Psi}$ with human annotations $\Psi$. The model predicts $M$ possible instances, denoted as $\hat{\Psi}{=}\{(\hat{\psi}^n, \hat{c}^n)\}_{n=1}^M$, sorted by their predicted relevance. Each instance $\hat{\psi}^n{=}(\hat{t}_s^n, \hat{t}_e^n)$ represents the beginning $(\hat{t}_s^n)$ and ending $(\hat{t}_e^n)$ times of the $n$-th relevant instance, and $\hat{c}^n$ is its corresponding confidence score. Specifically, $\hat{c}^n$ is a $k+1$ dimension vector when the query is an $k$ class action list, and it degenerates to dimension $2$ when the query is a descriptive language. 


\subsection{\Model~Architecture}
\label{sec:method-architecture}
As illustrated in Fig.~\ref{fig:arch}, our model pipeline includes three main modules, encoder, decoder, and denoiser.
First, we feed the video features $X_v$ into the encoder as a list of computation blocks. The query embeddings $X_l$ are concatenated with the video feature after a projection layer when it is available. We choose multi-head attention as our basic block unit due to its ability for long-range dependency modeling. Then, the output of the encoder, \ie memory, is forwarded to a decoder. The decoder is a stack of units that apply instance feature extraction and instance-instance interaction, where we show that the boundary information can be directly used to obtain effective proposal features. The boundary-denoising training (\ie, denoiser) module includes a new set of noisy spans to train the decoder, and it aims to help the decoder converges faster and better. The input to the denoiser is the memory from the encoder, and the noisy temporal spans are generated from the ground truth.
We disable the denoiser during inference as the ground truth spans are not available anymore. Also, we do not need NMS because a set loss was applied in training.

\subsubsection{Encoder}
Our encoder aims at modeling semantic information of the inputs, such as multi-modality interaction and non-local temporal dependency.
It includes the feature projection layer(s) followed by $N_{enc}$ transformer encoder layers. Given the video snippet features $X_v$, and optionally the language token features $X_l$, the feature project layer transforms its input into a joint space. Then we concatenate all the available features added with the sinusoidal positional embeddings to the video features, and send them to the transformer encoder layers. Since the self-attention mechanism has the potential to have message-passing over any two tokens, we assume the encoder output has rich semantic information over temporal relations and is conditioned by the given text prompt for video activity localization. Following the convention in DETR, we name this intermediate output as \textit{memory}.

\subsubsection{Decoder}
The decoder consists of $N_{dec}$ decoder layers, each of them takes the proposal start/end and corresponding proposal embeddings as inputs, and progressively refines the proposal boundaries and classifies their categories. Such $N_q$ proposals serve as \textit{queries} and interact with global video representation in the decoder, which is introduced in DETR. To avoid tendinous hand-crafted designs such as pre-defined anchors, the proposal start/end and proposal embeddings are set as learnable parameters and can be updated through backpropagation.

\begin{figure}[t]
\centering
\includegraphics[trim={0cm 0cm 0cm 0cm},width=.8\linewidth,clip,]{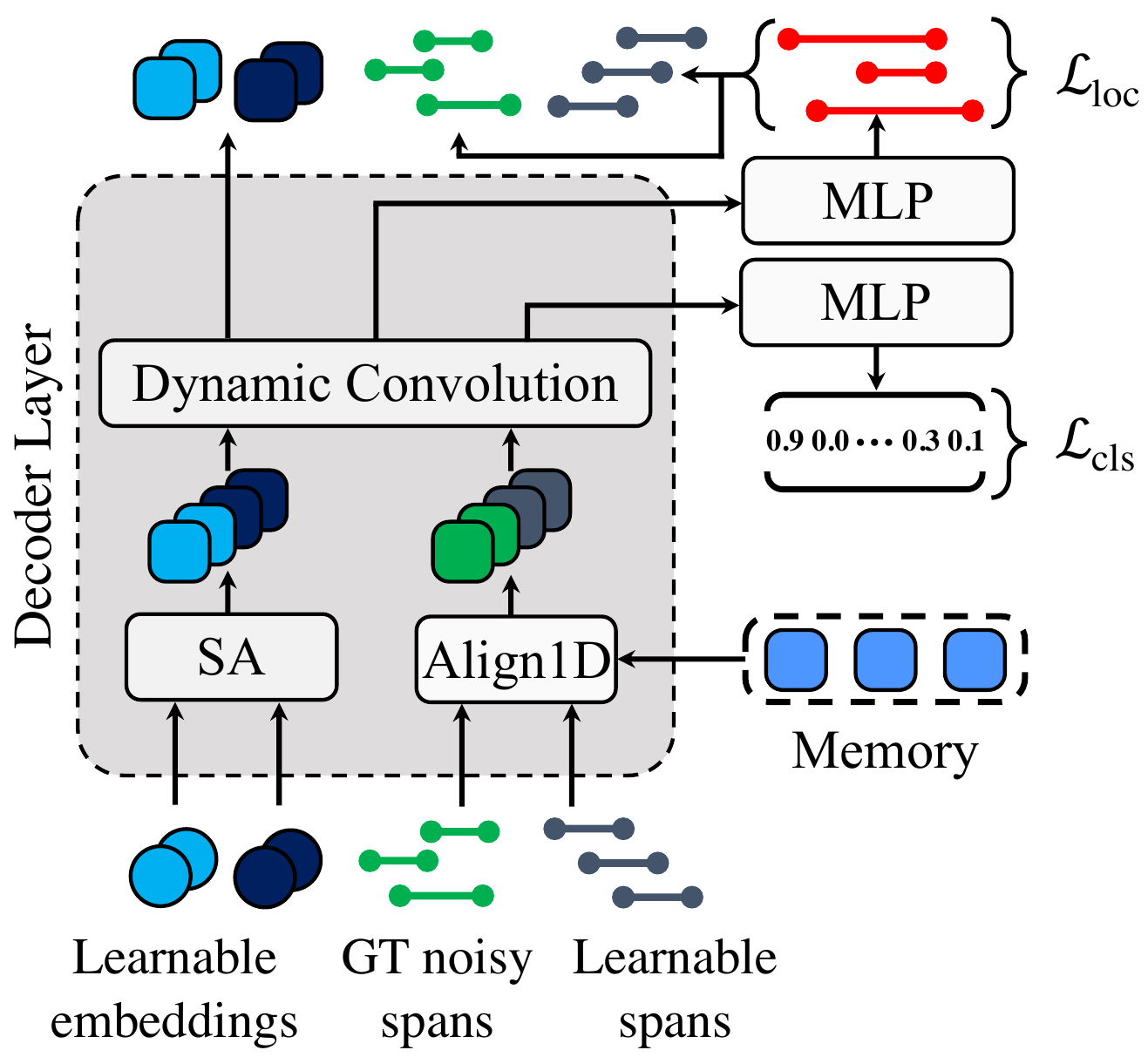}
\caption{\textbf{Decoder.} In each decoder layer, the self-attention is first adopted on the proposal embeddings to model the proposal relations with each other. Then the proposal features are explicitly extracted from the memory by RoI alignment, and they further interact with proposal embeddings to enrich the proposal representation through Dynamic Convolution. Last, the feed-forward MLPs updates proposal embeddings and proposal start/end gradually.
}
\label{fig:decoder}
\vspace{-0.2cm}
\end{figure}

In each decoder layer, the module follows the order of the self-attention layer, cross-attention layer, and feed-forward layer. To be specific, the self-attention layer is adopted on the proposal embeddings to model the proposal relations with each other. Next, the cross-attention layer simulates the interactions between the proposal embeddings with the encoder's output memory, thereby the proposal embeddings can capture the rich semantic information from the video and refine the proposal boundaries. Different from DETR, we find that explicitly modeling proposal features is crucial for proposal embedding interaction. Specifically, we replace the standard cross-attention layer in the transformer decoder with the DynamicConv layer, which is  proposed in \cite{sun2021sparse}. As illustrated in Fig.~\ref{fig:decoder}, temporal RoI alignment \cite{xu2020gtad} is first adopted to extract the proposal features based on the start/end locations, then the proposal features and proposal embeddings are sent into the DynamicConv module to accomplish the interactions between implicit proposal embeddings with explicit proposal features. The DynamicConv layer outputs the enhanced proposal features, and the updated proposal embeddings and updated proposal start/end locations are predicted with several feed-forward MLPs.

After each decoder layer, the proposal boundaries will be refined with predicted start/end offset. Together with the updated proposal embeddings, the refined proposals will be sent into the next decoder layer to regress more precise boundaries and more accurate classification probabilities.

\subsubsection{Boundary-denoising training}
\label{sec:method-train-inference}
The boundary-denoising training (denoiser) module  includes a new set of noisy spans to train the decoder. Therefore, it has similar input and output as designed in the decoder. Differently, the temporal spans in boundary-denoising training are randomly sampled around the ground truth activity location. These sampled spans are able to diversify the temporal inputs and accelerate model convergence. However, since the activity location is unknown in the test set, we disable this mechanism during model inference.

To be more specific, during training, given a set of ground-truth temporal activities, $\{(t_s^n, t_e^n)\}_{n=1}^{M_{gt}}$ and a noise level $\sigma$, we first sample $N_{qn}$ 2D vectors from Gaussian $\mathcal{N}(0,\sigma I)$, denoted as $\{\vec{\epsilon}_n\}_{n=1}^{N_{qn}}$, $\vec{\epsilon}_n \in \mathcal{R}^2$.
Then we divide the $N_{qn}$ noise vectors into two groups. The first group combines with ground truth action spans to create a positive set of proposals, \eg, one noisy proposal could be $(t_s^n+\epsilon_1^{n'}, t_e^n+\epsilon_2^{n'}), n\in\{1,\cdots,N_{gt}\},n'\in\{1,\cdots,N_{qn}\}$. The second group works as a negative set of proposals so that we directly use $(0.25+\epsilon_1^{n'}, 0.75+\epsilon_2^{n'}), n'\in\{1,\cdots,N_{qn}\}$ to make most of the proposals are valid, \ie, inside of the video. Noted that when a proposal has a boundary outside of $(0,1)$, we will clamp it into the valid range.

Once the noisy proposals are created, we feed them into the decoder to remove the noise in each decoder layer. Optionally, we can still have the learned proposal embedding and spans in the decoder, but the noisy proposal should not communicate with the learned proposal in case of annotation leakage. Moreover, the noisy proposals have their own assignment to the training target, and will not affect the matching algorithm applied in the learnable spans in the decoder. In inference, our predictions are from the learned proposal position, and the decoder has the capability to denoise the learned proposal in each layer.


\subsection{Loss functions}
We denote $\hat{\Psi}{=}\{(\hat{\psi}^n, \hat{c}^n)\}_{n=1}^M$ as the set of predictions, and $\Psi{=}\{\psi^n\}_{n=1}^{M_{gt}}$ as the set of ground-truth temporal activities, where $\hat{\psi}^n{=}(\hat{t}_s^n, \hat{t}_e^n)$ and $\psi^n{=}(t_s^n, t_e^n)$. To compute the loss, we need to first find an assignment $\pi$ between predictions and ground truth. We determine the assignment based on the matching cost $\mathcal{L}_{match}$ similar to DETR~\cite{carion2020end} as:
\begin{equation}
\mathcal{L}_{match}(\hat{\Psi}, \Psi) = \sum_{n=1}^{M_{gt}} [-\hat{c}^{\pi(n)} + \mathcal{L}_{span}(\hat{\psi}^{\pi(n)}, \psi^n)].
\end{equation}

With this matching cost, following previous works, we use the Hungarian algorithm \cite{kuhn1955hungarian} to find the optimal bipartite matching $\hat{\pi}$, where $\hat{\pi}{=}\argmin_{\pi} \mathcal{L}_{match}$. Based on this assignment $\hat{\pi}$, our overall loss is defined as:
\begin{equation}
    \mathcal{L} = \lambda_{loc} \mathcal{L}_{loc} + \lambda_{cls} \mathcal{L}_{cls} + \lambda_{sailency} \mathcal{L}_{sailency},
\end{equation}
where $\lambda_{loc}, \lambda_{cls}, \lambda_{sailency} \in \mathbb{R}$ are hyper-parameters balancing the three terms. $\mathcal{L}_{loc}$ combines an L1 loss and a generalized IoU loss (1D version as in \cite{rezatofighi2019generalized}) between the prediction and ground-truth to measure the localization quality:
\begin{equation}
    \mathcal{L}_{loc} = \lambda_{L_1}\mathcal{L}_1(\hat{\psi}^{\hat{\pi}(n)}, \psi^n)+ \lambda_{gIoU}\mathcal{L}_{gIoU}(\hat{\psi}^{\hat{\pi}(n)}, \psi^n),
\end{equation}
and $\mathcal{L}_{cls}$ is the cross entropy function to measure whether the action category is correctly classified or whether the prediction matches the query correctly. $\mathcal{L}_{sailency}$ directly borrows from Moment-DETR \cite{moment-detr} for query-dependent highlights detection. Hence, $\lambda_{sailency}$ is set to 0 for benchmarks without highlight detection.

%


\section{Experiments}
\label{sec:experiment}
This section presents our experimental setup, showcasing our pipeline's excellent performance. 
We first introduce datasets and metrics before reporting the comparison with the current state-of-the-art methods. We conclude the chapter with our thorough ablation study, which validates all our design choices. 


\begin{table*}[t]
\centering
\caption{\small{\bf Benchmarking of grounding methods on the MAD dataset.}  
We follow the methodology presented in~\cite{barrios2023localizing} and adopt a two-stage approach. In our work, we re-use the first stage implemented in~\cite{barrios2023localizing} and only work on the grounding (second stage) method. We 
report recall performance for baselines with ($\dagger$) and without the first-stage model.
}
\setlength{\tabcolsep}{2.5pt}
\renewcommand{\arraystretch}{1} 
\resizebox{\linewidth}{!}{%
\footnotesize
\begin{tabular}{@{}c@{\hspace{0.4em}} 
l 
c@{\hspace{0.2em}} 
ccccc c@{\hspace{0.2em}} 
ccccc c@{\hspace{0.2em}} 
cccc  c@{}}

\toprule
& \multirow{2}{*}{\textbf{Model}}
& \phantom{} & \multicolumn{5}{c}{\textbf{IoU=0.1}} 
& \phantom{} & \multicolumn{5}{c}{\textbf{IoU=0.3}}
& \phantom{} & \multicolumn{5}{c}{\textbf{IoU=0.5}} \\

\cmidrule{4-8} \cmidrule{10-14} \cmidrule{16-20} 
&
&& \textbf{R@1} & \textbf{R@5} & \textbf{R@10} & \textbf{R@50} & \textbf{R@100}
&& \textbf{R@1} & \textbf{R@5} & \textbf{R@10} & \textbf{R@50} & \textbf{R@100}
&& \textbf{R@1} & \textbf{R@5} & \textbf{R@10} & \textbf{R@50} & \textbf{R@100} \\
\midrule
          
&Zero-shot CLIP~\cite{Soldan_2022_CVPR}  
    && $6.57$ & $15.05$ & $20.26$ & $37.92$ & $47.73$ 
    && $3.13$ & $9.85$  & $14.13$ & $28.71$ & $36.98$ 
    && $1.39$ & $5.44$  &  $8.38$ & $18.80$ & $24.99$  \\
                                  
&VLG-Net~\cite{soldan2021vlg}     
    && $3.50$ & $11.74$ & $18.32$ & $38.41$ & $49.65$
    && $2.63$ & $9.49$  & $15.20$ & $33.68$ & $43.95$ 
    && $1.61$ & $6.23$  & $10.18$ & $25.33$ & $34.18$  \\
                                  
&Moment-DETR~\cite{moment-detr}   
    && $0.31$ & $1.52$ & $2.79$ & $11.08$ & $19.65$ 
    && $0.24$ & $1.14$ & $2.06$ & $7.97$  & $14.29$ 
    && $0.16$ & $0.68$ & $1.20$ & $4.71$  & $8.46$     \\

&\model~(ours)
    && $1.06$ & $4.07$ & $6.75$ & $20.07$ & $31.35$ 
    && $0.86$ & $3.34$ & $5.44$ & $15.67$ & $24.09$ 
    && $0.57$ & $2.17$ & $3.50$ & $9.73$  & $14.75$     \\

\cmidrule{1-20}
                                  
&$\dagger$Zero-shot CLIP~\cite{Soldan_2022_CVPR} && 
    $9.30$ & $18.96$ & $24.30$ & $39.79$ & $47.35$ 
    && $4.65$ & $13.06$ & $17.73$ & $32.23$ & $39.58$ 
    && $2.16$ & $7.40$ & $11.09$ & $23.21$ & $29.68$   \\

&$\dagger$VLG-Net~\cite{soldan2021vlg}    
    && $5.60$ & $16.07$ & $23.64$ & $45.35$ & $55.59$ 
    && $4.28$ & $13.14$ & $19.86$ & $39.77$ & $49.38$ 
    && $2.48$ & $8.78$ & $\underline{13.72}$ & $\mathbf{30.22}$ & $\mathbf{39.12}$  \\

&$\dagger$Moment-DETR~\cite{moment-detr} 
    && $5.07$ & $16.30$ & $24.79$ & $\underline{50.06}$ & $\underline{61.79}$ 
    && $3.82$ & $12.60$ & $19.43$ & $\underline{40.52}$ & $\underline{50.35}$ 
    && $2.39$ & $7.90$ & $12.06$ & $24.87$ & $30.81$  \\

&CONE~\cite{hou2022cone}
    && $\underline{8.90}$ & $\underline{20.51}$& $\underline{27.20}$ & $43.36$ & $-$ 
    && $\underline{6.87}$ & $\underline{16.11}$& $\underline{21.53}$ & $34.73$ & $-$ 
    && $\underline{4.10}$ & $\underline{9.59}$ & $12.82$ & $20.56$ & $-$     \\

&$\dagger$\model~(ours)
    && $\mathbf{11.59}$ & $\mathbf{30.35}$ & $\mathbf{41.44}$ & $\mathbf{66.07}$ & $\mathbf{73.62}$ 
    && $\mathbf{9.08}$  & $\mathbf{23.33}$ & $\mathbf{31.57}$ & $\mathbf{49.90}$ & $\mathbf{55.68}$ 
    && $\mathbf{5.63}$  & $\mathbf{14.03}$ & $\mathbf{18.69}$ & $\underline{29.12}$ & $\underline{32.51}$     \\
                                                        
\bottomrule
\end{tabular}%
}
\vspace{.1cm}
\label{tab:experiments-sota-mad}
\vspace{-.3cm}
\end{table*}

\subsection{Datasets}
\label{sec:experiment-datasets}
\noindent Unless otherwise specified, we adopt the original data splits for each dataset and report performance on the test set.

\noindent\textbf{MAD~\cite{Soldan_2022_CVPR}.} This recently released dataset comprises $384$K natural language queries (train $280{,}183$, validation $32{,}064$, test $72{,}044$) temporally grounded in $650$ full-length movies for a total of over $1.2$K hours of video, making it the largest dataset collected for the video language grounding task. Notably, it is the only dataset that allows the investigation of long-form grounding, thanks to the long videos it contains. 

\noindent\textbf{QVHighlights~\cite{lei2021detecting}.} This is the only trimmed video dataset for the grounding task, constituted by $10{,}148$ short videos with a duration of $150$s. Notably, this dataset is characterized by multiple moments associated with each query yielding a total of $18{,}367$ annotated moments and $10{,}310$ queries (train $7{,}218$, validation $1{,}550$, test $1{,}542$).

\noindent\textbf{TACoS~\cite{TACoS_ACL_2013}.} TACoS dataset consists of static camera videos of cooking activities. The dataset comprises only $127$ videos.  Each video contains an average of $148$ queries for a total of $18{,}818$ video-query pairs divided into three official splits (train $10{,}146$, validation $4{,}589$, test $4{,}083$).

\noindent\textbf{THUMOS-14~\cite{jiang2014thumos}.} The THUMOS dataset contains $413$ untrimmed videos annotated with $20$ action categories. Similar to previous works, we train over the $200$ videos in the validation split and test the model on the $213$ videos in the testing set. 


\subsection{Metrics}
\label{sec:experiment-metrics}

\noindent\textbf{Recall.} Our metric of choice for the grounding task is Recall@$K$ for IoU=$\theta$ (R@$K$-IoU=$\theta$). Given an ordered list of predictions, the metric measures if any of the top $K$ moments have a temporal IoU larger than $\theta$ with the ground truth span. Results are averaged across all samples. 

\noindent\textbf{Mean Average Precision (mAP).}  Following the literature, we compute the mAP metric with IoU thresholds $\{0.5, 0.7\}$.


\subsection{Implementation Details}
\label{sec:experiment-details}
Our implementation is built atop the Moment-DETR repository~\cite{lei2021detecting}, and we inherit their default settings unless stated otherwise. Our algorithm is compiled and tested using Python 3.8, PyTorch 1.13, and CUDA 11.6. Notably, we have increased the number of encoder and decoder layers to $8$, beyond which we observed saturation. We also use a fixed number of $30$ queries (proposals) during both training and inference.
To train our model, we use the AdamW~\cite{adamw} optimizer with a learning rate of $1e-4$ and weight decay of $1e-4$. We train the model for $200$ epochs and select the checkpoint with the best validation set performance for ablation. For evaluation, we compare our model with the state-of-the-art (SOTA) on the \textit{test} split. We do not use large-scale pre-training as in~\cite{liu2022umt}, but we train our models from random initialization. For detailed experimental settings on each dataset, please refer to the \textit{supplementary material}. To extract video and text features, we follow the methods described in the existing literature. Specifically, we use the MAD feature from the official release in~\cite{Soldan_2022_CVPR}, the QVHighlights feature from~\cite{lei2021detecting}, and the TACoS video and language feature from a pre-trained C3D~\cite{tran2015learning} and GloVe~\cite{pennington2014glove}, as described in~\cite{soldan2021vlg}. For the THUMOS-14 dataset, we use video features from TSN, as first used in~\cite{lin2018bsn}.


\begin{table}[t]
\centering
\caption{\small{\bf Benchmarking of grounding methods on the \textit{test} split of QVHighlights dataset.} }

\setlength{\tabcolsep}{2.5pt}
\renewcommand{\arraystretch}{1} 
\resizebox{\linewidth}{!}{%
\footnotesize
\begin{tabular}{@{}c@{\hspace{0.4em}} 
l 
c@{\hspace{0.2em}} 
cc c@{\hspace{0.2em}} 
ccc c@{}}

\toprule
& \multirow{2}{*}{\textbf{Model}}
& \phantom{} & \multicolumn{2}{c}{\textbf{R@1}} 
& \phantom{} & \multicolumn{3}{c}{\textbf{mAP}} \\

\cmidrule{3-5} \cmidrule{7-9}
&
&& \textbf{@0.5} & \textbf{@0.7} 
&& \textbf{@0.5} & \textbf{@0.75} &  \textbf{Avg.} \\

\midrule     
&MCN~\cite{hendricks17iccv}  
&& $11.41$ & $2.72$ && $24.94$ & $8.22$ & $10.67$ \\

&CAL~\cite{escorcia2019temporal}  
&& $25.49$ & $11.54$ && $23.40$ & $7.65$ & $9.89$ \\

&XML~\cite{lei2020tvr}  
&& $41.83$ & $30.35$ && $44.63$ & $31.73$ & $32.14$ \\

&XML+~\cite{lei2021detecting}  
&& $46.69$ & $33.46$ && $47.89$ & $34.67$ & $34.90$ \\

&Moment-DETR~\cite{moment-detr}   
&& $52.89$ & $33.02$ && $\underline{54.82}$ & $29.40$ & $30.73$ \\

&UMT~\cite{liu2022umt}           
&& $\underline{56.23}$ & $\underline{41.18}$ && $53.83$ & $\underline{37.01}$ & $\underline{36.12}$ \\

&\model~(ours)
&& $\mathbf{59.27}$ & $\mathbf{45.07}$ && $\mathbf{61.30}$ & $\mathbf{43.07}$ & $\mathbf{42.96}$ \\
                                            
\bottomrule
\end{tabular}%
}
\vspace{.1cm}
\label{tab:experiments-sota-qv}
\vspace{-.5cm}
\end{table}

\subsection{Comparison with State-of-the-Art}
\label{sec:experiment-sota}

Comparative studies are reported in this section. The best result is highlighted in bold in each table, while the runner-up is underlined. 

\noindent\textbf{MAD.} Tab.~\ref{tab:experiments-sota-mad} summarizes the performance of several baselines in the MAD dataset. For this evaluation, we follow the two-stage methodology introduced in~\cite{barrios2023localizing} and combine our \model~with a Guidance Model. This model serves as a first-stage method that conditions the scores of the second-stage grounding methods to boost their performance. We report the performance of the baselines with and without the Guidance Model, highlighting the former with the $\dagger$ symbol. Note that CONE~\cite{hou2022cone} does not use the same first stage, yet it implements the same concept with different details. 

First, let's analyze the second stage's performance (rows 1-4). We can observe that proposal-based methods such as Zero-Shot CLIP and VLG-Net~\cite{soldan2021vlg} offer much stronger performance with respect to the proposal-free approaches Moment-DETR~\cite{moment-detr} and \model. This observation is congruent with many other findings in the literature, even beyond the localization tasks video. In this case, the strongest challenge is constituted by the long-form nature of MAD, which naturally yields many false positive predictions. As it turns out, proposal-free methods are particularly susceptible to this phenomenon.
Nonetheless, the combination of guidance and grounding models is able to significantly boost these methods' recall by removing false-positive predictions. In fact, our \model~obtains the highest metric for most configurations, with large margins (between $9.6\%$ and $34.4\%$ relative improvements) against the runner-up methods. 

\noindent\textbf{QVHighlights.} We find that our method's good behavior translates to other datasets with very different characteristics, as in QVHighlights~\cite{lei2021detecting}. Recall and mAP performance are presented in Tab.~\ref{tab:experiments-sota-qv}, where \model~obtains the best performance for all metrics with relative improvements ranging from $3.4\%$ and $14.0\%$. Note, we strive for fair comparisons; therefore, we do not report Moment-DETR and UMT performance when pretraining is used. Additional details can be found in~\cite{liu2022umt}. 

\begin{table}[t]
\centering 
\setlength{\tabcolsep}{2pt}
\renewcommand{\arraystretch}{1} 
\caption{{\bf  Benchmarking of grounding methods on the TACoS dataset.}   }

\resizebox{\linewidth}{!}{
\footnotesize
\begin{tabular}{@{}c@{\hspace{0.4em}} 
l 
ccc c@{\hspace{0.2em}} 
ccc c@{}
}
\toprule
& \multirow{2}{*}{\textbf{Model}}
& \phantom{} & \multicolumn{3}{c}{\textbf{R@1}} 
& \phantom{} & \multicolumn{3}{c}{\textbf{R@5}} \\

\cmidrule{3-6} \cmidrule{8-10}
&
&& \textbf{@0.1} & \textbf{@0.3} & \textbf{@0.5} 
&& \textbf{@0.1} & \textbf{@0.3} & \textbf{@0.5} \\

\midrule
&MCN~\cite{Hendricks_2017_ICCV}       
&& $14.42$ & $-$     & $5.58$  && $37.35$ & $-$     & $10.33$  \\

&CTRL~\cite{Gao_2017_ICCV}            
&& $24.32$ & $18.32$ & $13.30$ && $48.73$ & $36.69$ & $25.42$  \\

&MCF~\cite{wu2018multi}               
&& $25.84$ & $18.64$ & $12.53$ && $52.96$ & $37.13$ & $24.73$  \\

&TGN~\cite{chen_etal_2018_temporally} 
&& $41.87$ & $21.77$ & $18.90$ && $53.40$ & $39.06$ & $31.02$  \\

&ACRN~\cite{ACRN_SIGIR_18}            
&& $24.22$ & $19.52$ & $14.62$ && $47.42$ & $34.97$ & $24.88$  \\

&ROLE~\cite{10.1145/3240508.3240549}  
&& $20.37$ & $15.38$ & $9.94$  && $45.45$ & $31.17$ & $20.13$  \\

&VAL~\cite{song2018val}               
&& $25.74$ & $19.76$ & $14.74$ && $51.87$ & $38.55$ & $26.52$  \\

&ACL-K~\cite{Ge_2019_WACV}            
&& $31.64$ & $24.17$ & $20.01$ && $57.85$ & $42.15$ & $30.66$  \\

&CMIN~\cite{lin2020moment}            
&& $36.68$ & $27.33$ & $19.57$ && $64.93$ & $43.35$ & $28.53$  \\

&SM-RL~\cite{wang2019language}        
&& $26.51$ & $20.25$ & $15.95$ && $50.01$ & $38.47$ & $27.84$  \\

&SLTA~\cite{jiang2019cross}           
&& $23.13$ & $17.07$ & $11.92$ && $46.52$ & $32.90$ & $20.86$  \\

&SAP~\cite{Chen_19_SAP}               
&& $31.15$ & $-$     & $18.24$ && $53.51$ & $-$     & $28.11$  \\

&TripNet~\cite{hahn2020tripping}      
&& $-$     & $23.95$ & $19.17$ && $-$     & $-$     & $-$      \\

&2D-TAN (P)~\cite{2DTAN_2020_AAAI}    
&& $47.59$ & $37.29$ & $25.32$ && $70.31$ & $57.81$ & $45.04$ \\

&DRN~\cite{Zeng_2020_CVPR}            
&& $-$     & $-$     & $23.17$ && $-$     & $-$     & $33.36$  \\

&CSMGAN~\cite{liu2020jointly}         
&& $42.74$ & $33.90$ & $27.09$ && $68.97$ & $53.98$ & $41.22$ \\

&GTR-H~\cite{cao-etal-2021-pursuit}
&& - & $40.39$ & $30.22$ && - & $61.94$ & $47.73$ \\


&IVG~\cite{nan2021interventional}
&& $49.36$ & $38.84$ & $29.07$ && $-$ & $-$ & $-$\\

&MSAT-2s~\cite{zhang2021multi}
&& $-$ & $44.56$ & $34.12$ && $-$ & $63.53$ & $\underline{53.59}$\\

&VLG-Net~\cite{soldan2021vlg}      
&&  $\underline{57.21}$ & $\underline{45.46}$ & $\underline{34.19}$ && $\mathbf{81.80}$ & $\mathbf{70.38}$ & $\mathbf{56.56}$ \\

&\model~(ours)
&&  $\mathbf{58.39}$ & $\mathbf{47.36}$ & $\mathbf{35.89}$ && $\underline{77.56}$ & $\underline{64.73}$ & $51.04$ \\
\bottomrule
\end{tabular}
}
\label{tab:experiments-sota-tacos}
\vspace{-0.5cm}
\end{table}
\noindent\textbf{TACoS.} In the TACoS benchmark, our \model~provides competitive performance that surpasses the current state of the art for R@1 $\forall \theta$. Yet, our approach falls short for R@5. We hypothesize that this shortcoming is to be imputed on the training strategy of proposal-free methods. In particular, we use the Hungarian matcher to associate the best prediction with the ground truth at training time. We would like to emphasize the advantage of \model~that only require very few queries (\ie, 30) compared with our competitors, as they usually have hundreds or thousands of proposals. This matcher, in turn, prevents us from exploiting the supervision provided by the other predictions, which are discarded toward the loss computation. The result of this strategy is to train for R@1 without addressing R@5, which can explain the results presented in Tab.~\ref{tab:experiments-sota-tacos}. 


\subsection{Ablations and Analysis}
\label{sec:experiment-ablations}
We design several ablations and analysis studies to probe our method and devise meaningful takeaways that can help in identifying fruitful future directions. 

\noindent\textbf{Noise level.}
Tab.~\ref{tab:ablation-noise} investigates the impact of varying noise levels (measured by noise-to-signal ratio). Specifically, we gradually increased the noise level from -20dB on the left to 20dB on the right to assess the model's performance. Our findings suggest that the model's performance follows a parabolic pattern as the noise level increases, with the best results achieved 44.66\% mAP in the (-10, 0] range. Additionally, the study also implies that a noise level that is too small will risk the model to overfit, resulting in unstable performance and higher prediction variance. Conversely, a noise level that is too large will also lead to underfiting, reducing prediction variance but causing a corresponding decrease in model accuracy.

\begin{table}[t]
\small
\caption{\textbf{Ablation study on noise ratio.} A small noise can cause the model to underfit, while a large noise may reduce the prediction variance and also decrease the performance.  }
\centering
\resizebox{\linewidth}{!}{
\footnotesize
\begin{tabular}{c|cccc}
\toprule
\textbf{Noise Ratio (dB)} & \textbf{(-20,-10]} & \textbf{(-10,0]} & \textbf{(0,10]} & \textbf{(10,20]} \\
\midrule
Avg. mAP  (avg)      & $43.71$     & $44.66$   & $44.22$  & $42.89$ \\
Avg. mAP  (std)      & $~~1.64$     & $~~0.86$   & $~~0.56$  & $~~0.40$ \\
\toprule
\end{tabular}
}
\label{tab:ablation-noise}
\vspace{-0.2cm}
\end{table}
\begin{table}[t]
\small
\caption{\textbf{Ablation study on our model architecture on the \textit{val} split of the VQ-highlight dataset.} When self-attention is applied, we observe significant improvements, which suggests the proposal-proposal interaction plays important role in global message passing. Also, nearly all the metrics got improved once the noisy temporal spans participated in model training.}
\centering
\resizebox{\linewidth}{!}{
\footnotesize
\begin{tabular}{cccccc}
\toprule
\multicolumn{1}{c}{\multirow{1}{*}{\textbf{\makecell{Dynamic \\ Conv}}}} & \multicolumn{1}{c}{\multirow{1}{*}{\textbf{\makecell{Self \\ Att.}}}} & \multicolumn{1}{c}{\multirow{1}{*}{\textbf{\makecell{Denoise \\ Training}}}} & \multicolumn{2}{c}{\textbf{R@1}}       & \multicolumn{1}{c}{\textbf{mAP}}                                                  \\ 
\cline{4-6} 
& &           & \multicolumn{1}{c}{\textbf{@0.5}} & \multicolumn{1}{c}{\textbf{@0.7}} & \multicolumn{1}{c}{\textbf{Avg.}} \\ 
\midrule
\multicolumn{3}{c}{baseline (Moment-Detr) } & $53.94$ & $34.84$ & $32.20$
\\ 
\midrule
\xmark                                  & \xmark                                  & \xmark                                  & $54.13$                        & $38.97$                       & $32.30$                     \\
\cmark                                  & \xmark                                  & \xmark                                  & $46.06$                       & $34.19$                        & $28.59$                    \\
\xmark                                  & \cmark                                  & \xmark                                  & $57.23$                       & $\mathbf{44.97}$                      & $41.08$                    \\
\cmark                                  & \cmark                                  & \xmark                                  & $\mathbf{58.13}$                        & $44.65$                        & $\mathbf{42.95}$                  \\ \midrule
\xmark                                  & \xmark                                  & \cmark                                  & $57.16$                       & $44.13$                       & $38.47$                     \\
\cmark                                  & \xmark                                  & \cmark                                  & 55.87                     & $43.42$                  & $41.63$                  \\
\xmark                                  & \cmark                                  & \cmark                                 & $57.81$                       & $45.29$                      & $40.76$                         \\
\cmark                                  & \cmark                                  & \cmark                                 & $\mathbf{59.87}$                      & $\mathbf{45.87}$                     & $\mathbf{44.56}$
\\
\bottomrule
\end{tabular}
}
\label{tab:ablation-arch}
\end{table}

\noindent\textbf{Decoder design.}
We first explore the best way to incorporate the proposal span information in the decoder in the second block of Tab.~\ref{tab:ablation-arch}. Then in the last block of this table, we study the model behavior when denoising training is applied. 
\textbf{1.} Our model is built upon moment-DETR, a transformable encoder-decoder network for moment localization. 
Row 2 and Row 3 compare different ways to combine the proposal feature with the query feature in the moment-DETR framework, but the self-attention module is disabled. We found when Dynamic Conv. is not applied (Row 3), which means we use the average of the feature and proposal feature to represent the action instance, the model shows improved performance. 
\textbf{2.} Row 4 and Row 5 show when self-attention is applied, we observe significant improvements, which suggests the proposal-proposal interaction plays important role in global message passing. 
\textbf{3.} In the last block, \ie Row 6-9, nearly all the metrics got improved once the noisy temporal spans participated in model training.  The improvements over non-self-attention models (Row 6, 7) are more significant, although we don't introduce any extra computation in the decoder. We also find the best model performance is achieved by enabling all the modules, especially Dynamic Conv. This is because the Dynamic Conv. operation is more sensitive to the temporal feature of the proposals, and our denoising training, by jittering the proposal boundaries, can bring more augmentations in the target data domain.

\begin{table}[t]
\small
\caption{\textbf{Ablation study on the number of queries.} When we have fewer queries, adding noise helps the model quickly converge to a stable state, resulting in improved performance. However, increasing the query number will slow down the convergence rate because more negatives are included.} 
\centering
\resizebox{\linewidth}{!}{
\begin{tabular}{cclllll}
\toprule
\multirow{2}{*}{\textbf{\makecell{Query \\ Number}}} & \multirow{2}{*}{\textbf{\makecell{Denoise \\ Training}}} & \multicolumn{5}{c}{\textbf{Epoch Number}}           \\
\cline{3-7} 
                         &                           & \multicolumn{1}{c}{\multirow{1}{*}{\textbf{5}}}     & \multicolumn{1}{c}{\multirow{1}{*}{\textbf{10}}}     & \multicolumn{1}{c}{\multirow{1}{*}{\textbf{20}}}     & \multicolumn{1}{c}{\multirow{1}{*}{\textbf{40}}}     & \multicolumn{1}{c}{\multirow{1}{*}{\textbf{80}}}     \\ \midrule
\multirow{2}{*}{8}      & \xmark                    & $12.60$ & $26.30$ & $36.44$ & $39.65$ & $41.46$ \\
                         & \cmark                    & $11.84$ & $30.62$ & $39.73$ & $41.59$ & $42.18$ \\ \midrule
\multirow{2}{*}{16}      & \xmark                    & $4.92$ & $26.80$ & $38.35$ & $41.82$ & $42.22$ \\
                         & \cmark                    & $12.26$ & $31.45$ & $39.34$ & $42.52$ & $43.78$ \\ \midrule
\multirow{2}{*}{32}      & \xmark                    & $7.91$ & $25.37$ & $39.07$ & $41.38$ & $42.27$ \\
                         & \cmark                    & $4.16$ & $22.36$ & $38.42$ & $43.98$ & $44.11$ \\
                         \bottomrule
\end{tabular}
}\label{tab:ablation-query}
\vspace{-0.3cm}
\end{table}

\noindent\textbf{Query number.} Tab.~\ref{tab:ablation-query} presents the performance comparison of models trained on different epochs with varying numbers of queries. The first two blocks suggest that after adding noise, the model can quickly converge to a stable state, resulting in improved performance. This observation is consistent with the results shown in Fig.~\ref{fig:pool}. Increasing the number of queries from 8 to 16 and then to 32, adding noise at the beginning helps the model to converge faster and perform better. However, as the number of queries exceeds a certain threshold, the convergence speed slows down, and performance deteriorates. This may be due to the increasing number of negative samples resulting from one-to-one assignments performed by the Hungarian matcher, which makes it harder for the model to learn how to denoise and causes decreased performance. This finding also highlights that our model can achieve excellent performance with a small number of predictions, rendering NMS unnecessary.

\begin{table}[bt]
\caption{\textbf{Action localization results on test set of THUMOS14}, measured by mAP (\%) at different tIoU thresholds.}
\small
\resizebox{\linewidth}{!}{
\begin{tabular}{lccccc}
	\toprule
	\textbf{Method}          & \textbf{0.3}  & \textbf{0.4}  & \textbf{0.5}  & \textbf{0.6}  & \textbf{0.7} \\
	\midrule
    BMN~\cite{lin2019bmn}                  & $56.0$ & $47.4$ & $38.8$ & $29.7$ & $20.5$    \\
    G-TAD~\cite{xu2020g}                  & $57.3$ & $51.3$ & $43.0$ & $32.6$ & $22.8$    \\
    VSGN~\cite{zhao2021video}                 & $66.7$ & $60.4$ & $52.4$ & $41.0$ & $30.4$    \\
    AFSD~\cite{lin2021learning}               & $67.3$ & $62.4$ & $55.5$ & $43.7$ & $31.1$    \\
    E2E-TAL~\cite{liu2022empirical}                      & $69.4$ & $64.3$ & $56.0$ & $46.4$ & $34.9$ \\
    DaoTAD~\cite{wang2021rgb}    & $72.7$ & - & $59.8$ & - & $33.3$ \\
    TALLFormer~\cite{cheng2022tallformer}      & $76.0$ & - & $63.2$ & - & $34.5$ \\
    \midrule
    \textbf{Ours w.o. Denoise}                & $75.64$ & $70.42$ & $62.62$ & $\mathbf{51.36}$ & $38.58$ \\
    \textbf{Ours w. Denoise}                & $\mathbf{76.65}$ & $\mathbf{72.06}$ & $\mathbf{64.26}$ & ${51.23}$ & $\mathbf{39.54}$ \\
    \bottomrule
\end{tabular}
}
\label{tab:sota_thu}
\vspace{-0.3cm}
\end{table}

\begin{table}[b]
\vspace{-.2cm}
\small
\caption{\textbf{Applying our denoising model into a diffusion framework.} We don't observe any evident performance gain of DiffusionLoc from DenoiseLoc, showing a strong efficiency and effectiveness of our method
in temporal activity localization.}
\centering
\resizebox{\linewidth}{!}{
\begin{tabular}{r|cccccc}
\toprule
\textbf{Steps} & \textbf{1}     & \textbf{2}     & \textbf{4}     & \textbf{8}     & \textbf{64}    & \textbf{512}   \\
\midrule
Avg. mAP           & $\mathbf{44.26}$ & $43.85$ & $43.04$ & $43.61$ & $43.57$ & $42.99$ \\
mAP@0.5             & $\mathbf{62.08}$ & $61.08$ & $60.21$ & $60.89$ & $60.66$ & $59.71$ \\
mAP@0.7             & $\mathbf{44.96}$ & $43.65$ & $43.04$ & $44.74$ & $44.35$ & $43.20$ \\
\bottomrule
\end{tabular}
}
\label{tab:study-diffusion}
\end{table}

\noindent\textbf{Generalizability on temporal action detection.} We also test our methods for the temporal action detection tasks on THUMOS dataset, by removing the text token in the encoder. As shown in Fig.~\ref{tab:sota_thu}, our methods without denoising training already achieves state-of-the-art detection performance. Adding denoise training helps the model converge, and the mAP@0.5 is further improved from 62.62\% to 64.26\%.


\subsection{Further Discussion}
\label{sec:discussion}
\noindent\textbf{DenoiseLoc \vs DiffusionLoc.} Our model can be viewed as a conditional diffusion model when time embedding (parameter governing the diffusion process) is also included in the dynamic convolutional layer. In this case, the input of our model is a temporal span of a certain level of noise, and the output is the offset of the proposal to the target activity localization. 
As an extension of Tab.~\ref{tab:ablation-noise}, we design a DDPM~\cite{ho2020denoising}-like training protocol~\cite{chen2022diffusiondet} to optimize our model, dubbed as DiffusionLoc. Specifically, in each training iteration, we random sample an integer $t$ in $\{1,\cdots,T\}$, and decide the noise level $\sigma$ based on it. Then the following process is the same as DenoiseLoc, we denoise the ground truth action location to train the network, and prediction the denoised proposal in model inference. We set $T$ in ${1,2,4,8,64,512}$ and report the model performance in Tab.~\ref{tab:study-diffusion}.

Surprisingly, we don't observe any evident performance gain of DiffusionLoc from DenoiseLoc, which achieves $44.66\pm0.86$ on the same validation split. One potential reason is that knowing the noise level does not help the model to localize activities more precisely, and denoising training is already an efficient and effective method in temporal activity localization. 

\subsection{Visualizations}
\label{sec:experiment-visualization}
\begin{figure}[t]
    \vspace{-0.1cm}
    \centering
    \includegraphics[trim={3.5cm 4cm 8cm .7cm},width=\linewidth,clip]{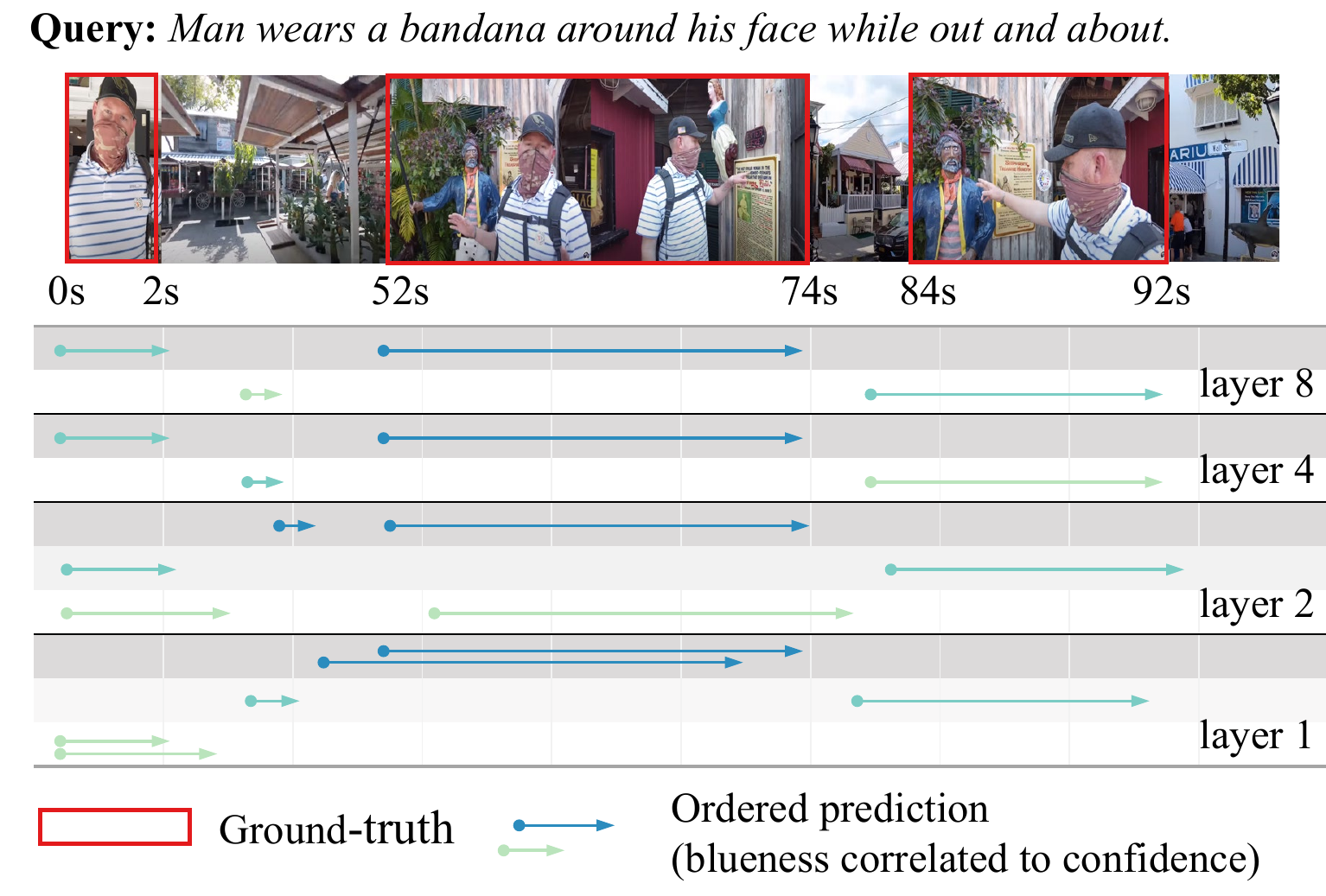}
    \caption{\textbf{Qualitative Results on QV-Highlights.} A deeper color means the proposal's confidence score is higher. Our method progressively denoises the action boundaries and removes the redundancies. See Sec.~\ref{sec:experiment-visualization} for more details. Best viewed in color. } 
    \label{fig:vis}
\end{figure}

We visualize the predictions from decoder layers 1, 2, 4, and 8 of a video example in Fig.~\ref{fig:vis}. We use colored arrows to denote predicted temporal spans (the darker the color the higher the model confidence). We also filtered the predictions if the confidence is less than $0.2$. As
the decoder goes deeper (bottom to up), the proposal boundaries
become more precise, and the ranking confidence scores become
more accurate. In the meantime, the redundant proposals are also suppressed. 

\section{Conclusion}
\label{sec:conclusion}
We propose DenoiseLoc, an encoder-decoder model,
which introduces a novel boundary-denoising training
paradigm to address the challenge of uncertain action boundaries in video activity localization. 
DenoiseLoc captures the relations within and across modalities in the encoder and progressively refines learnable proposals and noisy ground truth spans in multiple decoder layers. Our boundary-denoising training jitters action proposals and serves as an augmentation to guide the model on predicting meaningful boundaries under the uncertainty of initial noisy spans. Extensive experiments to demonstrate the effectiveness of DenoiseLoc, achieving state-of-the-art performance on several datasets, including the large-scale MAD dataset.

\setcounter{section}{0}
\renewcommand\thesection{\Alph{section}}
\section{Experiment settings}
This section presents the detailed experimental setup for each dataset evaluated in the main paper. Please also see `code.zip' for our implementation and `opt.json' for all the hyper-parameters. 

\noindent\textbf{MAD~\cite{Soldan_2022_CVPR}.} We used this dataset to evaluate our method under the long-form video setting. To have a fair comparison with the literature, we follow a coarse-to-fine two-stage method. We only use the DenoiseLoc prediction as a finer prediction in a short window that is predicted by~\cite{barrios2023localizing}. Also, we reduce the number of queries from 30 to 14 and do not find any performance drop. In addition, since MAD does not have snippet-wise annotation that could be used as saliency loss in Eq.~(2), we randomly sample 3 snippets inside the temporal span annotation as positives to help the encoder's convergence. 

\noindent\textbf{QVHighlights~\cite{lei2021detecting}.} Our training process follows precisely the description of our Sec.~4.3. To receive the test set performance, we submit our prediction to the official server hosted on \href{https://codalab.lisn.upsaclay.fr/competitions/6937}{CodaLab}. We hide our rank on the leaderboard to be anonymous, but the prediction file can be verified in the supplementary materials, named as ``qv\_submission.zip''.

\noindent\textbf{TACoS~\cite{TACoS_ACL_2013}.} This traditional dataset also does not have snippet-wise annotation for saliency loss in Eq.~(2), so we sample positive snippets as on the MAD dataset. We also find the start/end loss, commonly used (\eg,~\cite{sun2021sparse,soldan2021vlg}) on this dataset, is also helpful to find the global optima. Therefore, we have two extra loss terms for snippet-level supervision.

\noindent\textbf{THUMOS-14~\cite{jiang2014thumos}.} We disable the language encoding part because it is unavailable. Also, we use the binary classifier with focal loss to predict $c$ action categories. No more video label fusion is applied after the prediction. 

\section{Experiment on egocentric video dataset}
Due to space limitations, we move our experiments of the egocentric video dataset, Ego4D~\cite{ego4d}, into the supplementary materials. Ego4D is a massive-scale, egocentric dataset for human daily activity recordings. Among the 3,670 hours of video, we validate our algorithm on the split of Natural language queries (NLQ). This split has 227-hour videos; each query consists of a video clip and a query expressed in natural language. The target is to localize the temporal window span within the video history where the answer to the question can be found.

We follow the experiment settings as in ReLER~\cite{https://doi.org/10.48550/arxiv.2207.00383} and compare our method with the solutions from other challenge participants in Tab.~\ref{tab:nlq}. Note that our experiment and comparison are based on the initial annotation NLQ version 1.0, while the recently released version 2.0 was right before the submission deadline. Tab.~\ref{tab:nlq} clearly shows that our method surpasses others on top-1 recall over IoU 0.3 and 0.5 without model ensemble. However, our method falls behind on the top-5 recall, which is consistent with Tab.~3 for the TACoS experiment. We hypothesize this is because the Hungarian matcher associates the only best prediction with the ground truth at training time, and we only use a few queries (\ie, 30).
\begin{table}[t]
\centering 
\caption{{\bf  Benchmarking of grounding methods on the Ego4D dataset for NLQ task (\textit{test} split).} We clearly surpass others on top-1 recall over IoU 0.3 and 0.5 without model ensemble. }
\vspace{1mm}
\small
\begin{tabular}{
l 
c@{\hspace{0.2em}}
cc 
c@{\hspace{0.2em}}   
cc
}
\toprule
\textbf{Model}
\phantom{} && \multicolumn{2}{c}{\textbf{IoU=0.3(\%)}} 
\phantom{} && \multicolumn{2}{c}{\textbf{IoU=0.5(\%)}} \\

\cmidrule{3-4} 
\cmidrule{6-7}

&& \textbf{R@1} & \textbf{R@5} 
&& \textbf{R@1} & \textbf{R@5}  \\

\midrule
VSLNet~\cite{ego4d} 
&& 5.47& 11.21 && 2.80 & 6.57 \\

2D-TAN~\cite{ego4d} 
&& 5.80& 13.90 && 2.34 & 5.96 \\

ReLER~\cite{https://doi.org/10.48550/arxiv.2207.00383} 
&& 12.89& 15.41 && 8.14 & 9.94 \\

EgoVLP~\cite{egovlp} 
&& 10.46& 16.76 && 6.24 & 11.29\\

CONE~\cite{hou2022cone} 
&& 15.26& \textbf{26.42} && 9.24 & \textbf{16.51}\\

InternVideo~\cite{https://doi.org/10.48550/arxiv.2211.09529} 
&& 16.45 & 22.95  &&  10.06& 16.10 \\

\model~(ours) 
&& \textbf{19.33}& 21.48 && \textbf{11.94} & 13.89\\

\bottomrule
\end{tabular}
\label{tab:nlq}
\end{table}

\section{Denoise training on decoder (denoiser)}
\noindent
\textbf{Training as a denoiser}. The proposed denoising training can be viewed in two ways. (1) \textit{Augmented training set.} Besides the learned proposal span, we generate random spans around the ground-truth location. Those new spans can accelerate the decoder's convergence and make our training more robust. (2) \textit{Individual module.} Since the generated spans and learned spans do not interact in the transformer decoder layers, they can be studied as in two different modules independently, while the two modules share the same model architecture and weights. Therefore, the individual module, \eg denoiser, works similarly to a regularizer.

\noindent
\textbf{Spans in inference}.
Although the denoiser module is disabled during inference as the ground-truth information is not available anymore, we discover that replacing the learnable spans with random spans also gives a competitive performance, see Tab.~\ref{tab:appendix_rnd_learned}. 
The average performance (\eg AmAP in \textit{all} setting) stays almost the same, but using random spans gives more advantages under a more strict metric IoU=0.7 and is also good at middle/short activities, shown in the last two columns. Therefore, we can deduce the learned temporal spans are good at long activities over lenient constraints, while random spans have a higher chance of predicting precise action boundaries over short activities. 
\begin{table}[t]
\small
\caption{\textbf{Comparison of different spans during inference.} Replacing the learnable spans with
random ones also give competitive performance.} %
\centering
\begin{tabular}{ccccccc}
\toprule
\multirow{2}{*}{span} & \multicolumn{2}{c}{Top 1 Recall} & \multicolumn{4}{c}{Average mAP}  \\ \cline{2-3} \cline{4-7}
                      & @0.5            & @0.7           & all    & long   & middle & short \\ \midrule
learned               & \textbf{60.52}          & 45.23        & 44.41 & \textbf{52.67} & 45.10  & 11.68 \\
random                 & 59.29          & \textbf{46.84}        & \textbf{44.51} & 50.82 & \textbf{45.75} & \textbf{12.10} \\
\bottomrule
\end{tabular}\label{tab:appendix_rnd_learned}
\end{table}

\begin{figure}[ht]
\centering
\includegraphics[trim={.3cm 0cm 1.1cm 0cm},width=.95\linewidth,clip,]{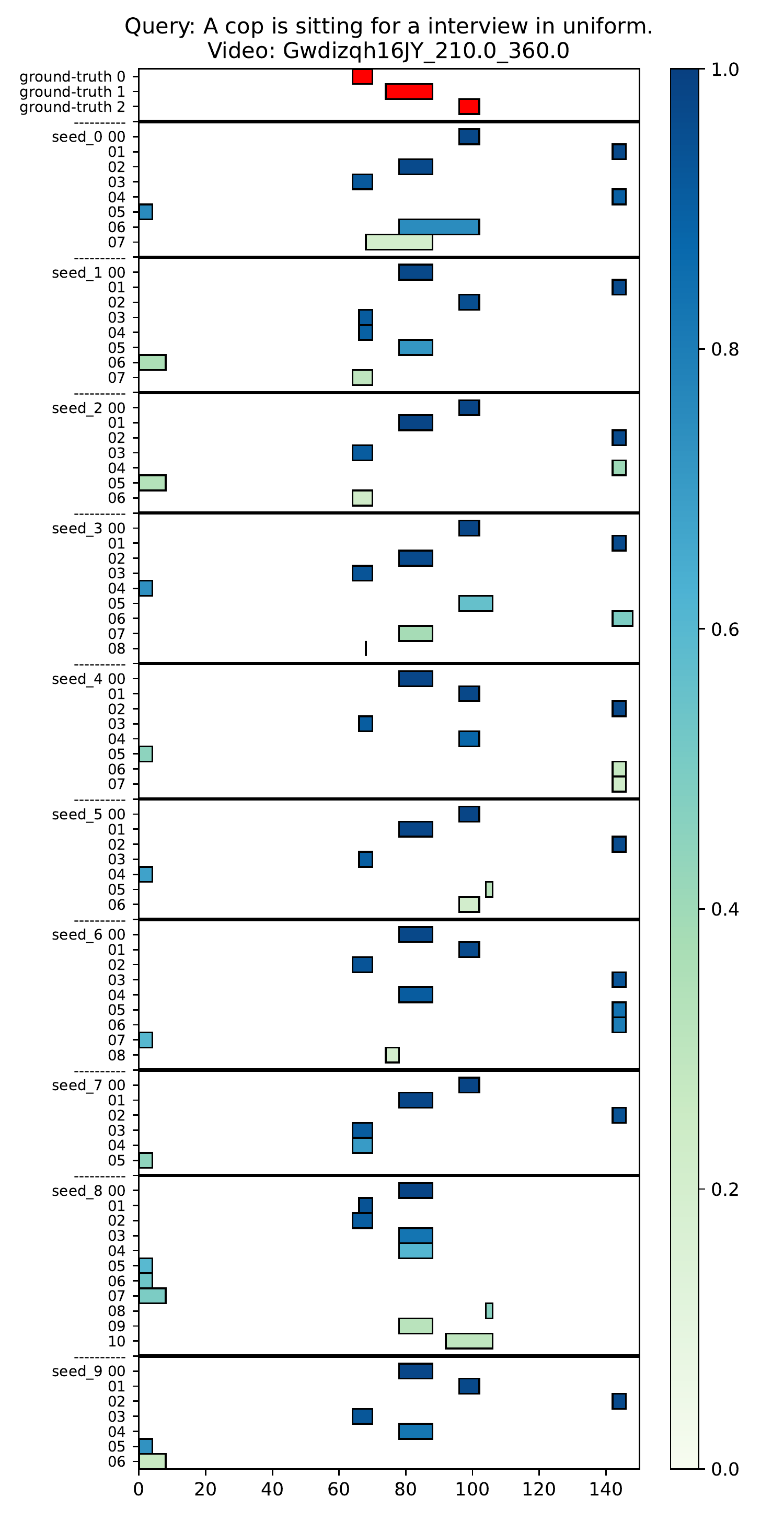}
\caption{\textbf{Visualization with ten random seeds.} Almost all subplots have correct predictions, although orders are different. \href{https://youtu.be/Gwdizqh16JY?t=270}{Link.}}
\label{fig:appendix_10254}
\end{figure}

\begin{figure}[ht]
\centering
\includegraphics[trim={.3cm 0cm 1.1cm 0cm},width=.95\linewidth,clip,]{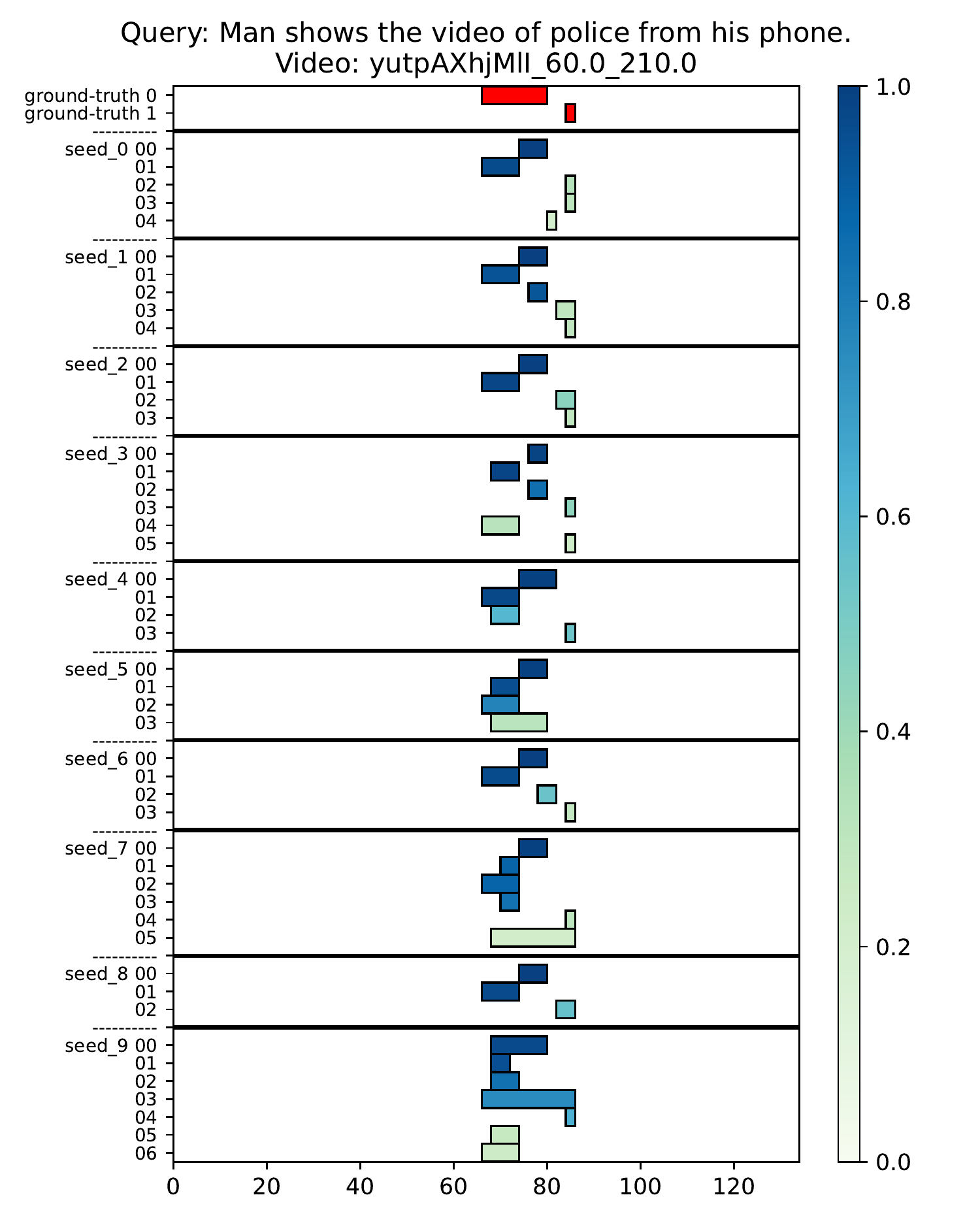}
\caption{\textbf{Visualization with ten difference random seeds.} The model is more sensitive to the random seed in this case. \href{https://youtu.be/yutpAXhjMlI?t=130}{Link.}}
\label{fig:appendix_10046}
\end{figure}

\noindent
\textbf{Uncertainty in inference}.
Following the random span setting in inference, another good property of DenoiseLoc is that the uncertainty of model predictions can be measured by generating random spans with different random seeds. To explore this, we change the seed from $0$ to $9$, run inference, and visualize the predictions on the same video from all runs in Fig.~\ref{fig:appendix_10254} and Fig.~\ref{fig:appendix_10046}. The ground-truth spans in red are in the top subplot, and the predictions from ten different seeds are in the bottom. In each subplot of the predictions, the one with higher confidence is higher and of darker color. Predictions with confidence scores less than $0.2$ are removed. The unit on the x-axis is per second. 

Fig.~\ref{fig:appendix_10254} visualizes our results to the query ``\textit{A cop is sitting for an interview in uniform.}'' We observe that almost all subplots have correct top-3 predictions, although their order could differ. Experiment with seed 8 misses a target probability because all the random spans are far away from this location. Thus we don't find any more predictions after timestamp 110 (x-axis) for seed 8. 

Fig.~\ref{fig:appendix_10046} is more challenging according to our predictions because the model is more sensitive to the random seed. We observe both the central location and the duration of the top-1 prediction vary a lot over the 10 runs, while it seems that only the seed-9 experiment gives a correct answer. However, if we take a closer look at the \href{https://youtu.be/yutpAXhjMlI?t=130}{video}, there is a \textit{camera-change} from 2:13-2:15 (equivalent to 73-75 in our x-axis). Therefore, our model predicts it as two separate activities, but both are relevant to the query, ``\textit{Man shows the video of police from his phone.}'' 

\section{Supplementary to our diffusion variant}

\noindent
\textbf{DiffusionLoc}
A time embedding can be incorporated into the dynamic convolutional layer in the decoder. We name this variant of our model as DiffusionLoc. To be more concrete, given an integer $t$, we project its sinusoidal position embeddings by multi-layer perception to produce scale and shift vectors, denoted as $s_1$ and $s_2$. Given the original output $X$ after the FC layer, our time-conditioned output is $(s_1+1)X+s_2$. Note that we have used $\psi$ as spans, so the decoder can be formulated as $$\psi_{pred}=Dec( \psi_{noise}^{(t)}, t, Memory, \Theta)$$, where $Memory$ is the output from the encoder and $\Theta$ is the learnable parameters of the model.

The training phase of DiffusionLoc is pretty much the same as DenoiseLoc, while the difference is that the noise level is controlled by $t$, randomly sampled in $\{1,2,\cdots,T\}$ in each iteration and each video, \ie, $$\psi_{noise}^{(t)}=\sqrt{\bar{\alpha_t}}\psi_{gt}+\sqrt{1-\bar{\alpha_t}}\vec{\epsilon},$$
where $T=1000$, the choice of $\bar{\alpha_t}$ follows DDPM~\cite{ho2020denoising}, and $\vec{\epsilon}$ is the noise term as defined in our DenoiseLoc.

The ideal inference of DiffusionLoc is to take from $T, T-1, \cdots$ until $1$. In each step, we estimate the noisy spans at the noise level at $t$, and the decoder will predict the activity intervals from the estimated spans. Mathematically, each step can be shown as,
\begin{align*}
\psi_{pred}^{(t)}&=Dec( \tilde{\psi}_{noise}^{(t)}, Memory, t) \\ \tilde{\psi}_{noise}^{(t)}&=\sqrt{\bar{\alpha_t}}\psi_{pred}^{(t+1)}+\sqrt{1-\bar{\alpha_t}}\vec{\epsilon}.
\end{align*}
In practice, we observe the predicted negative spans degenerate quickly and cannot be recovered to a proper random distribution, while the predicted positive spans also have the risk of being negative after the \textit{diffusion} process. Therefore, it is best for us to have only one step in inference. 
However, given a converged DenosieLoc model, we can always progressively boost the model performance by using different sets of initial spans and ensemble the predictions from each run. We would like to gain attention from the community and leave this study and follow-up works from DenoiseLoc.

\section{Limitations and Ethics Concerns}

\noindent
\textbf{Limitations:} A common limitation of deep learning-based computer vision solutions for video understanding is the potential for overfitting. This is driven by the adoption of very large models and sometimes small datasets. In this work, we address this concern by evaluating the benefit of our approach across multiple datasets, some of which are of massive scales, like MAD and Ego4D. 
Yet, our work presents itself with other limitations. In fact, throughout the experimental section, we mentioned how our performance for high recall (i.e., R@5) might not achieve state-of-the-art results (example in Tab.~\ref{tab:nlq}). We hypothesized that this phenomenon is due to the Hungarian matcher that is adopted in this work and several others (\cite{Tan_2021_RTD}). Such an algorithm assigns the ground truth to the best prediction, allowing us to compute the loss only for R@1 and not optimize for multiple predictions. We believe researchers should focus on improving this aspect of the current set-prediction approaches. We leave the solution to this limitation to future works. 

\noindent
\textbf{Ethics Concerns:} As the performance of video understanding algorithms gradually improves, such techniques, publicly available for research purposes, might be adopted for illicit surveillance purposes, raising concerns about privacy and potential misuse. Our work is released to the public in accordance with the limitation of the MIT license. Any misuse of the information provided in this document is to be condemned and reprehended.

{\small
\bibliographystyle{ieee_fullname}
\bibliography{arxiv}
}

\end{document}